\documentclass[11pt,a4paper]{article}
\usepackage[hyperref]{eacl2021}
\usepackage{graphicx}

\usepackage{times}
\usepackage{latexsym}

\usepackage{microtype}

\usepackage{multirow}
\usepackage{graphicx}
\usepackage{bm}
\usepackage[whole]{bxcjkjatype}
\usepackage{arydshln}
\usepackage{amsmath,amssymb,mathrsfs}
\usepackage{enumitem}

\usepackage{natbib}

\makeatletter
\newcommand{\tblcaption}[1]{\def\@captype{table}\caption{#1}}

\aclfinalcopy 

\makeatletter
\def\Hline{
  \noalign{\ifnum0=`}\fi\hrule \@height 3.\arrayrulewidth \futurelet
  \reserved@a\@xhline}
\makeatother

\title{Understanding Pre-Editing for Black-Box Neural Machine Translation}

\newcommand{\affiA}{{$^{\dagger}$}}
\newcommand{\affiB}{{$^{\ddagger}$}}

\author{Rei Miyata{\affiA} \qquad Atsushi Fujita{\affiB}\vspace{5pt}\\
  {\affiA}Nagoya University\\
  Furo-cho, Chikusa-ku, Nagoya, 464-8601, Japan\\
  \texttt{miyata@nuee.nagoya-u.ac.jp}\\
  {\affiB}National Institute of Information and Communications Technology\\
  3-5 Hikaridai, Seika-cho, Souraku-gun, Kyoto, 619-0289, Japan\\
  \texttt{atsushi.fujita@nict.go.jp}}
\date{}

\begin{document}
\maketitle
\begin{abstract}
Pre-editing is the process of modifying the source text (ST) so that it can be translated by machine translation (MT) in a better quality. Despite the unpredictability of black-box neural MT (NMT), pre-editing has been deployed in various practical MT use cases. Although many studies have demonstrated the effectiveness of pre-editing methods for particular settings, thus far, a deep understanding of what pre-editing is and how it works for black-box NMT is lacking. To elicit such understanding, we extensively investigated human pre-editing practices. We first implemented a protocol to incrementally record the minimum edits for each ST and collected 6,652 instances of pre-editing across three translation directions, two MT systems, and four text domains. We then analysed the instances from three perspectives: the characteristics of the pre-edited ST, the diversity of pre-editing operations, and the impact of the pre-editing operations on NMT outputs. Our findings include the following: (1) enhancing the explicitness of the meaning of an ST and its syntactic structure is more important for obtaining better translations than making the ST shorter and simpler, and (2) although the impact of pre-editing on NMT is generally unpredictable, there are some tendencies of changes in the NMT outputs depending on the editing operation types.
\end{abstract}

\section{Introduction}
\label{sec:intro}

Recent advances in machine translation (MT) have greatly facilitated its practical use in various settings from business documentation to personal communication. In many practical cases, MT systems are used as black-box and one well-tested approach to make use of a black-box MT is pre-editing, i.e., modifying the source text (ST) to make it suitable for the intended MT system.

The effectiveness of pre-editing has so far been demonstrated in many studies \citep{ASLIB-1990-Pym,MTS-2007-O'Brien,LREC-2014-Seretan}. A study focusing on statistical MT (SMT) has also shown that more than 90\% of an ST can be rewritten into a text that can be machine-translated with sufficient quality \citep{EAMT-2017-Miyata}, exhibiting the potential of the pre-editing approach.

However, the feasibility and possibility of pre-editing for neural MT (NMT) has not been examined extensively. While efforts have recently been invested in the implementation of pre-editing strategies for black-box NMT settings, achieving improved MT quality \citep[e.g.,][]{MTS-2019-Hiraoka,AAAI-2020-Mehta}, the potential gains of pre-editing remain unexplored. Notably, the impact of pre-editing on black-box MT is unpredictable in nature. In particular, NMT models trained in an end-to-end manner can be sensitive to minor modifications of the ST \citep{ACL-2019-Cheng}, which may affect the feasibility of pre-editing.

In short, while pre-editing has been implemented in practical MT use cases, what pre-editing is and how it works with black-box NMT systems remain open questions. To explore the possibility of pre-editing and its automation, in this study, we provide fine-grained analyses of human pre-editing practices and their impact on NMT. We systematically collected pre-editing instances in various conditions, i.e., translation directions, NMT systems, and text domains (\S\hyperlink{sec-collection}{3}). We then conducted in-depth analyses of the collected instances from the following three perspectives: the characteristics of the pre-edited ST (\S\hyperlink{sec-characteristics}{4}), the diversity of pre-editing operations (\S\hyperlink{sec-diversity}{5}), and the impact of pre-editing operations on the NMT outputs (\S\hyperlink{sec-impact}{6}). The findings of these analyses provide useful insights into the effective and efficient implementation of pre-editing for the better use of black-box NMT systems in the future, as well as the robustness of current NMT systems when STs are manually perturbed.

\section{Related Work}
\label{sec:related-work}
Pre-editing is the process of rewriting the source text (ST) to be translated in order to obtain better translations by MT. Though the scope of effective pre-editing operations depends on the downstream MT system and there is no deterministic relation between pre-editing operations and the quality of MT output, its effectiveness has been demonstrated for various translation directions, MT architectures, and text domains.

Manual pre-editing has long been implemented in combination with controlled languages \citep{ASLIB-1990-Pym,CLAW-2003-Reuther,Benjamins-2003-Nyberg,CL-2014-Kuhn}. In the period of rule-based MT (RBMT), pre-editing was considered as a promising approach since the behaviour of RBMT is more predictable and controllable. For example, \citet{MTS-2007-O'Brien} examined the impact of English controlled language rules on two different MT engines, revealing the rules of high effectiveness. The pre-editing approach with controlled languages has also been tested for statistical MT (SMT) \citep{MTS-2007-Aikawa,EAMT-2012-Hartley,LREC-2014-Seretan}. These studies developed or utilised a set of controlled language rules for rewriting ST. While these rule sets are optimised for particular MT systems and differ from each other, we can observe some shared characteristics among them. In particular, rules that prohibit long sentences (e.g., of more than 25 words) are widely adopted in the existing rule sets \citep{CLAW-2003-O'Brien}.

Automation of pre-editing is also an important research field in natural language processing. Semi-automatic tools such as controlled language checkers \citep{MT-2001-Bernth,CLAW-2003-Mitamura} and interactive rewriting assistants \citep{ACL-2013-Mirkin,AIETI-2015-Gulati} were developed to facilitate manual pre-editing activities. Fully automatic pre-editing has long been explored \citep[e.g.,][]{CLAW-1998-Shirai,NLPRS-2001-Mitamura,MT-2001-Yoshimi,EAMT-2010-Sun}. In particular, many researchers have examined methods of \emph{reordering} the source-side word order as a pre-translation processing \citep{COLING-2004-Xia,ACL-2007-Li,ACL-2015-Hoshino}. While the reordering approach has generally proven effective for SMT, its effectiveness for NMT is not obvious; negative effects have even be reported \citep{WAT-2015-Zhu,PBML-2017-Du}. In recent years, techniques of automatic text simplification have been applied to improve NMT outputs \citep{ATA-2018-Stajner,AAAI-2020-Mehta}. The underlying assumption of these studies is that simpler sentences are more machine translatable.

Previous studies have investigated various pre-editing methods from different perspectives, focusing on different linguistic phenomena. Indeed, individual research has led to improved MT results. However, what is crucially needed is a broad understanding of what pre-editing is and how it works. For example, \citet{EAMT-2017-Miyata} addressed this issue by collecting instances of \emph{bilingual pre-editing}, i.e., pre-editing ST while referring to its MT output, done by human editors and analysing them in detail. They demonstrated the maximum gain of pre-editing for an SMT and provided a comprehensive typology of editing operations. Nevertheless, their study has two major limitations: (1) recent NMT was not examined, and (2) practical insights for better practices of pre-editing were not sufficiently presented. 

\begin{table*}[!t]
\small
\centering
\begin{tabular}{lp{12.3cm}} \hline
\multirow{3}{*}{\textbf{5. Perfect}} & Information in the original text
has been completely translated. There are no grammatical errors in the
translation. The word choice and phrasing are natural even from a native
speaker's point of view. \\ \hline
\multirow{2}{*}{\textbf{4. Good}} & The word choice and phrasing are slightly
unnatural, but the information in the original text has been
completely translated, and there are no grammatical errors in the
translation. \\ \hline
\multirow{2}{*}{\textbf{3. Fair}} & There are some minor errors in the
translation of less important information in the original text, but
the meaning of the original text can be easily understood. \\ \hline
\multirow{2}{*}{\textbf{2. Acceptable}} & Important parts of the original
text are omitted or incorrectly translated, but the core meaning of
the original text can still be understood with some effort. \\ \hline
\textbf{1. Incorrect/nonsense} & The meaning of the original text is
incomprehensible. \\ \hline
\end{tabular}
\caption{MT evaluation criterion adopted in \citet{EAMT-2017-Miyata}: The ``Perfect'' and ``Good'' ratings are regarded as satisfactory quality.}
\label{tab:criteria}
\end{table*}

\begin{figure*}[!t]
  \def\@captype{table}
  \begin{minipage}[t]{.32\textwidth}
    \centering
    \includegraphics[width=5cm]{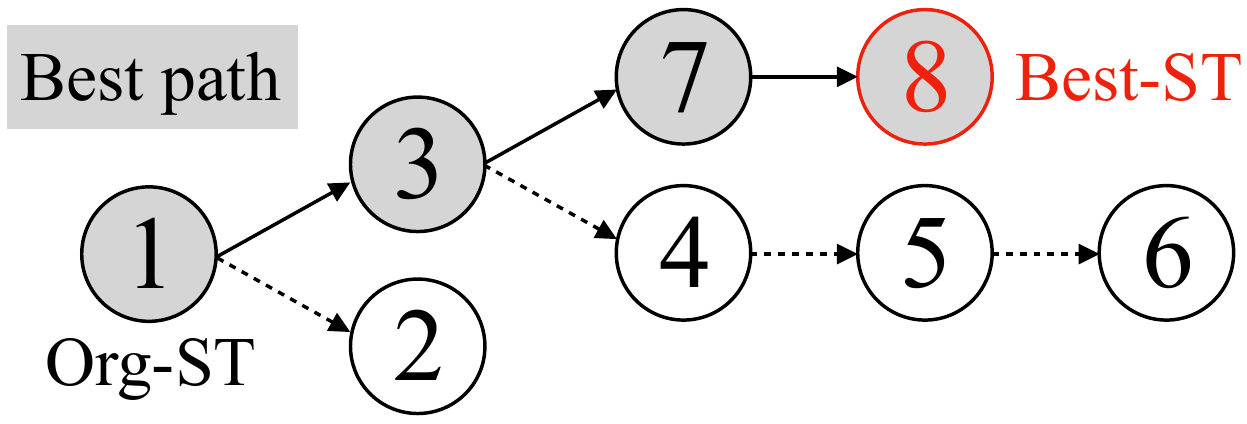}
    \caption{Tree representation of ST versions in a unit.}
    \label{fig:best-path}
  \end{minipage}
  \hfill
  \begin{minipage}[c]{.65\textwidth}
\small
\centering
\begin{tabular}{lp{2.7cm}ccc}
\hline
\textbf{Name} & \multicolumn{1}{c}{\textbf{Domain}} & \textbf{Mode} & \textbf{Size} & \multicolumn{1}{c}{\textbf{Avg.  length (S.D.)}} \\ \Hline
hospital & hospital conversation & spoken & 25 & 13.0 ~(4.7)~ \\ \hline
municipal & municipal procedure & written & 25 & 20.4 (10.7) \\ \hline
\multirow{2}{*}{bccwj} & Japanese-origin news article from BCCWJ & \multirow{2}{*}{written} & \multirow{2}{*}{25} & \multirow{2}{*}{28.6 (18.6)} \\ \hline
\multirow{2}{*}{reuters} & English-origin news article from Reuters & \multirow{2}{*}{written} & \multirow{2}{*}{25} & \multirow{2}{*}{36.8 (15.3)} \\ \hline
\end{tabular}
    \tblcaption{Statistics for the Org-ST datasets for pre-editing.}
    \label{tab:dataset}
  \end{minipage}
\end{figure*}

NMT models trained in an end-to-end manner behave very differently from SMT and RBMT, which, in turn, affects pre-editing practices. As reported in several studies, despite their rapid improvement, NMT models are still vulnerable to input noise \citep{ICLR-2018-Belinkov,COLING-2018-Ebrahimi,ACL-2019-Cheng,ACL-2020-Niu}. The pre-editing operations identified in previous studies are not necessarily effective for current black-box NMT systems.\footnote{The ideal goal of the pre-editing approach is to adapt the STs to what the intended NMT system can properly translate, and in the end, what it has been trained on, i.e., training data.  For a black-box MT system, because we cannot directly refer to its training data, we should grasp its statistical characteristics indirectly through MT output.} For example, \citet{MT-2019-Marzouk} adopted nine controlled language rules\footnote{The rules are as follows: (1) using straight quotes for interface texts, (2) avoiding light-verb construction, (3) formulating conditions as if sentences, (4) using unambiguous pronominal references, (5) avoiding participial constructions, (6) avoiding passives, (7) avoiding constructions with ``sein'' + ``zu'' + infinitive, (8) avoiding superfluous prefixes, and (9) avoiding omitting parts of the words \citep[p.184]{MT-2019-Marzouk}.} and evaluated their impact on the MT output for German-to-English translation in the technical domain. The human evaluation results revealed that these rules improved the performance of the RBMT, SMT, and hybrid systems, but did not have positive effects on the NMT system. \citet{MTS-2019-Hiraoka} demonstrated the effectiveness of the following three pre-editing rules in improving Japanese-to-English TED Talk subtitle translation using a black-box NMT system: (1) inserting punctuation, (2) making implied subjects and objects explicit, and (3) writing proper nouns in the target language (English).

As these studies cover a limited range of linguistic phenomena, translation directions, and text domains, we are not in the position to draw decisive conclusions; we still do not know what types of pre-editing operations are possible and how NMT is affected when these operations are performed. To elicit the best pre-editing practices for NMT, as a starting point, we need to understand what is happening and what can be obtained in the process of pre-editing, while also re-examining the previous findings and conventional methods.

\hypertarget{sec-collection}{\section{Collection of Pre-Editing Instances}}
\label{sec:collection}

\hypertarget{sub-protocol}{\subsection{Protocol}}
\label{sub:protocol}

To collect fine-grained manual pre-editing instances, we adopted the protocol formalised by \citet{EAMT-2017-Miyata}, in which a human editor incrementally and minimally rewrites an ST on a trial-and-error basis with the aim of obtaining better MT output. An original ST (Org-ST) and its pre-edited versions are collectively called a \emph{unit}. Using an online editing platform we developed, editors implement the protocol in the following steps:

\begin{description} 
  \setlength{\parskip}{0cm} 
  \setlength{\itemsep}{0cm} 
  \item[Step 1.] Evaluate the MT output of the current ST based on a 5-point scale criterion shown in Table~\ref{tab:criteria}. If the quality of the MT output is satisfactory (i.e., ``Perfect'' or ``Good''), go to Step 4; otherwise, go to Step 2.
  \item[Step 2.] Select one of the versions of the ST in the unit to be rewritten and go to Step 3. If none of the versions are likely to become satisfactory through further edits, go to Step 4.
  \item[Step 3.] Minimally edit the ST\footnote{We operationally defined ``to minimally edit'' as ``to modify an ST with a small edit that is difficult to be further decomposed into more than one independent edit, without inducing ungrammaticality in the edited sentence.''} while maintaining its meaning, referring to the corresponding MT output. The MT output for the edited ST is automatically generated and registered in the unit. Return to Step 1.
  \item[Step 4.] Select one version of the ST that achieves the best MT quality (Best-ST) from among all the versions in the unit, and terminate the process for the unit. 
\end{description}

The pre-editing instances in a unit collected through this protocol form a tree structure as shown in Figure~\ref{fig:best-path}.
We refer to the shortest path between the Org-ST and the Best-ST as \emph{Best path}. An important extension to the work in \citet{EAMT-2017-Miyata} is that our platform provides editors with a visualisation of the tree representation of the pre-editing history. This can facilitate the selection of ST versions in Step 2.

\begin{table*}[t]
\small
\centering
\begin{tabular}{llllrrrrlcc}
\hline
\multirow{2}{*}{\textbf{Lang.}} & \multirow{2}{*}{\textbf{System}} & \multirow{2}{*}{\textbf{Domain}} &  & \multicolumn{4}{l}{\textbf{Num. of pre-editing instances}} &  & \multicolumn{2}{c}{\textbf{Num. of units}} \\ \cline{5-8} \cline{10-11} 
 &  &  &  & \multicolumn{1}{l}{\textbf{Total}} & \multicolumn{1}{l}{\textbf{Avg.}} & \multicolumn{1}{l}{\textbf{Med.}} & \multicolumn{1}{l}{\textbf{Max}} &  & \multicolumn{1}{c}{\textbf{Org=Satisfactory}} & \multicolumn{1}{c}{\textbf{Best=Satisfactory}} \\ \Hline
\multirow{8}{*}{\textbf{Ja-En}} & \multirow{4}{*}{\textbf{Google}} & \textbf{hospital} &  & 255 & 10.2 & 7 & 55 &  & 4/25  & \textbf{25/25} \\
 &  & \textbf{municipal} &  & 162 & 6.5 & 5 & 44 &  & 9/25  & \textbf{25/25} \\
 &  & \textbf{bccwj} &  & 545 & 21.8 & 10.5 & 171 &  & 7/25  & 23/25  \\
 &  & \textbf{reuters} &  & 370 & 14.8 & 6.5 & 80 &  & 7/25  & \textbf{25/25} \\ \cline{2-11} 
 & \multirow{4}{*}{\textbf{TexTra}} & \textbf{hospital} &  & 139 & 5.6 & 5.5 & 25 &  & 7/25  & \textbf{25/25} \\
 &  & \textbf{municipal} &  & 136 & 5.4 & 4 & 35 &  & 10/25  & \textbf{25/25} \\
 &  & \textbf{bccwj} &  & 493 & 19.7 & 11.5 & 79 &  & 2/25  & 22/25  \\
 &  & \textbf{reuters} &  & 492 & 19.7 & 18 & 86 &  & 4/25  & 24/25  \\ \hline
\multirow{8}{*}{\textbf{Ja-Zh}} & \multirow{4}{*}{\textbf{Google}} & \textbf{hospital} &  & 264 & 10.6 & 10 & 30 &  & 0/25 & 24/25  \\
 &  & \textbf{municipal} &  & 376 & 15.0 & 13 & 41 &  & 0/25 & 23/25  \\
 &  & \textbf{bccwj} &  & 427 & 17.1 & 16 & 41 &  & 2/25  & 20/25  \\
 &  & \textbf{reuters} &  & 304 & 12.2 & 10 & 27 &  & 0/25 & 24/25  \\ \cline{2-11} 
 & \multirow{4}{*}{\textbf{TexTra}} & \textbf{hospital} &  & 160 & 6.4 & 6.5 & 15 &  & 1/25  & \textbf{25/25} \\
 &  & \textbf{municipal} &  & 172 & 6.9 & 7 & 20 &  & 2/25  & \textbf{25/25} \\
 &  & \textbf{bccwj} &  & 231 & 9.2 & 5 & 38 &  & 4/25  & 22/25  \\
 &  & \textbf{reuters} &  & 249 & 10.0 & 7 & 31 &  & 1/25  & 22/25  \\ \hline
\multirow{8}{*}{\textbf{Ja-Ko}} & \multirow{4}{*}{\textbf{Google}} & \textbf{hospital} &  & 209 & 8.4 & 9 & 22 &  & 0/25 & \textbf{25/25} \\
 &  & \textbf{municipal} &  & 225 & 9.0 & 8 & 26 &  & 0/25 & \textbf{25/25} \\
 &  & \textbf{bccwj} &  & 223 & 8.9 & 7 & 27 &  & 1/25  & 22/25  \\
 &  & \textbf{reuters} &  & 293 & 11.7 & 10 & 33 &  & 0/25 & 24/25  \\ \cline{2-11} 
 & \multirow{4}{*}{\textbf{TexTra}} & \textbf{hospital} &  & 160 & 6.4 & 6 & 26 &  & 2/25  & \textbf{25/25} \\
 &  & \textbf{municipal} &  & 171 & 6.8 & 5 & 32 &  & 2/25  & \textbf{25/25} \\
 &  & \textbf{bccwj} &  & 277 & 11.1 & 6 & 28 &  & 3/25  & 23/25  \\
 &  & \textbf{reuters} &  & 319 & 12.8 & 11 & 38 &  & 1/25  & 23/25  \\ \hline
\end{tabular}
\caption{Statistics for the collected pre-editing instances and the MT quality achievement.}
\label{tab:collected-instances}
\end{table*}

\hypertarget{sub-implementation}{\subsection{Implementation}}
\label{sub:implementation}

To extensively investigate pre-editing phenomena, we prepared the following conditions:
\begin{description} 
  \setlength{\parskip}{0cm} 
  \setlength{\itemsep}{0cm} 
  \item [Translation directions:] We targeted Japanese-to-English (Ja-En), Japanese-to-Chinese (Ja-Zh), and Japanese-to-Korean (Ja-Ko) translations.
  \item [MT systems:] As black-box MT systems, we adopted Google Translate\footnote{https://translate.google.com/} and TexTra.\footnote{https://textra.nict.go.jp/} Both are general-purpose NMT systems that are prevalently used for translating Japanese texts into other languages.
  \item [Text domains:] We selected four text domains, whose linguistic characteristics, such as mode and sentence length, are different from each other (see Table~\ref{tab:dataset} for details).
\end{description}

We randomly selected 25 Japanese sentences for each of the four text domains, and used the resulting ST set consisting of 100 sentences for all of the six combinations of translation direction and MT system. We assigned one editor to each translation direction. Each editor was asked to work with both MT systems, without being informed of the type of MT system used in the task. All editors were professional translators with sufficient writing skills in Japanese and experience for evaluating MT outputs. Before the commencement of the formal tasks, we trained the editors using example sentences so that they could become accustomed to the task and platform.

The Ja-En task was implemented from November to December 2019; the Ja-Zh and Ja-Ko tasks were implemented from December 2019 to February 2020.

\hypertarget{statistics}{\subsection{Statistics}}
\label{sub:statistics}

Table~\ref{tab:collected-instances} shows statistics for the pre-editing instances collected through the protocol described above. In general, the numbers of collected instances for the \textbf{hospital} and \textbf{municipal} domains were smaller than those for the \textbf{bccwj} and \textbf{reuters} domains, reflecting the influence of sentence length of the Org-ST. In other words, the shorter the sentence is, the fewer parts there are to be edited.

A notable finding is that while only about 11\% (69/600) of the MT output for the Org-ST was of satisfactory quality, 95\% (571/600) of the MT output of the Best-ST was satisfactory. This means that almost all the ST can be pre-edited into a form that can lead to satisfactory MT output, demonstrating the potential of both pre-editing and NMT.

The number of collected instances can be interpreted as the editing efforts required to obtain the Best-ST from the Org-ST. In most of the settings, the median number of collected instances for a unit falls in the range of 5 to 10. It is thus necessary to optimise the pre-editing process for an intended MT system. The length of the Best path approximates the minimum editing efforts needed to obtain the Best-ST. The total number of pre-editing instances in the Best path was 2,443, while the total of all instances is 6,652. This implies that there is substantial opportunity for reduction of the pre-editing efforts.

\begin{table*}[t]
\setlength{\tabcolsep}{1.95mm}
\small
\centering
\begin{tabular}{p{2.9cm}lcp{-0.3cm}cclcclcc}
\hline
 &  & \multicolumn{1}{c}{\multirow{2}{*}{\textbf{Org-ST}}} &  & \multicolumn{2}{c}{\textbf{Ja-En (Best-ST)}} &  & \multicolumn{2}{c}{\textbf{Ja-Zh (Best-ST)}} &  & \multicolumn{2}{c}{\textbf{Ja-Ko (Best-ST)}} \\ \cline{5-6} \cline{8-9} \cline{11-12} 
 &  & \multicolumn{1}{c}{} &  & \multicolumn{1}{c}{\textbf{Google}} & \multicolumn{1}{c}{\textbf{TexTra}} &  & \multicolumn{1}{c}{\textbf{Google}} & \multicolumn{1}{c}{\textbf{TexTra}} &  & \multicolumn{1}{c}{\textbf{Google}} & \multicolumn{1}{c}{\textbf{TexTra}} \\ \Hline
\textbf{Sentence length} & \textbf{Avg.} & 25.4 &  & 27.8 & 26.9 &  & 28.6 & 27.1 &  & 27.8 & 26.9 \\
\textbf{} & \textbf{S.D.} & 16.3 &  & 17.6 & 16.7 &  & 17.2 & 16.0 &  & 16.7 & 16.6 \\
\textbf{} & \textbf{Med.} & 19.5 &  & 21.5 & 20 &  & 23 & 22 &  & 22.5 & 20.5 \\ \hline
\textbf{Attachment distance} & \textbf{Avg.} & 1.95 &  & 1.97 & 1.99 &  & 1.99 & 1.99 &  & 2.00 & 1.98 \\
\textbf{(Avg. per sentence)} & \textbf{S.D.} & 0.65 &  & 0.53 & 0.65 &  & 0.60 & 0.63 &  & 0.64 & 0.62 \\
\textbf{} & \textbf{Med.} & 1.83 &  & 2.00 & 1.96 &  & 2.00 & 1.98 &  & 2.00 & 1.91 \\ \hline
\textbf{Dependency depth} & \textbf{Avg.} & 3.57 &  & 3.73 & 3.68 &  & 3.73 & 3.77 &  & 3.78 & 3.76 \\
\textbf{} & \textbf{S.D.} & 1.91 &  & 1.97 & 1.88 &  & 1.89 & 1.93 &  & 1.85 & 1.92 \\
\textbf{} & \textbf{Med.} & 3 &  & 3 & 3 &  & 3 & 4 &  & 4 & 4 \\ \Hline
\textbf{Lexical diversity} & \textbf{Token (A)}\hspace{-5pt} & 2,538 &  & 2,779 & 2,685 &  & 2,861 & 2,709 &  & 2,780 & 2,693 \\
 & \textbf{Type (B)} & 1,010 &  & 1,074 & 1,060 &  & 1,106 & 1,061 &  & 1,068 & 1,055 \\
 & \textbf{A/B} & 2.513 &  & 2.588 & 2.533 &  & 2.587 & 2.553 &  & 2.603 & 2.553 \\ \hline
\textbf{Word frequency rank} & \textbf{25th} & 7 &  & 7 & 7 &  & 7 & 7 &  & 7 & 7 \\ 
\textbf{(Percentile)} & \textbf{50th (Med.)} & 170 &  & 143 & 154 &  & 143 & 155 &  & 143 & 169.5 \\ 
 & \textbf{75th} & 2655 &  & 2304.25 & 2458 &  & 2471 & 2554 &  & 2470 & 2593.5 \\ 
\hline
\end{tabular}
\caption{Linguistic characteristics of the Org-ST and Best-ST.}
\label{tab:characteristics}
\end{table*}

\hypertarget{sec-characteristics}{\section{Characteristics of Pre-Edited Sentences}}
\label{sec:characteristics}

To understand the differences between the original and pre-edited STs, in this section, we describe their general linguistic characteristics. Here, we compare the Org-ST and the Best-ST that achieved a satisfactory MT result in order to elicit the features of machine translatable ST.

\hypertarget{sub-structural-characteristics}{\subsection{Structural Characteristics}}
\label{sub:structural-characteristics}

To quantify structural complexity, we used the following three indices:
\begin{enumerate}[label=(\arabic*)]
  \setlength{\parskip}{0cm} 
  \setlength{\itemsep}{0cm} 
\item sentence length: the number of words per sentence\footnote{If ST instance includes multiple sentences, we averaged the scores.}
\item attachment distance: the averaged distance of all attachment pairs of the Japanese base phrases in a sentence
\item dependency depth: the maximum distance from the root word in the dependency tree
\end{enumerate}
We used the Japanese tokeniser MeCab\footnote{https://taku910.github.io/mecab/} to calculate (1) and the Japanese dependency parser JUMAN/KNP\footnote{http://nlp.ist.i.kyoto-u.ac.jp/index.php?KNP} to calculate (2) and (3). 

The first three blocks in Table~\ref{tab:characteristics} show the results for these indices. It is evident on all indices, the Org-ST exhibits the lowest scores. In other words, the length and surface complexity of the sentences generally increased through the pre-editing operations. This is a counter-intuitive finding in that most previous pre-editing practices have axiomatically assumed that shorter and less complex sentences are better for MT. We further delve into this in \S\hyperlink{sub-typology}{5}.

\hypertarget{sub-lexical-characteristics}{\subsection{Lexical Characteristics}}
\label{sub:lexical-characteristics}

The remaining two blocks in Table~\ref{tab:characteristics} present statistics for the lexical characteristics of the STs. The results for lexical diversity indicate that both the total number of word types and the Token/Type ratio increased from the Org-ST to the Best-ST for all the conditions. This suggests that though the diversity of words increased slightly, the word distribution became peakier through pre-editing.

We also calculated the word frequency rank with Wikipedia as the reference.\footnote{We used the whole text data of Japanese Wikipedia obtained in October 2019 (https://dumps.wikimedia.org/).} To assess the status of word frequency in relation to MT, it would be ideal to use the training data for each MT system, but such data are unavailable in black-box MT settings. Therefore, we decided to use Wikipedia as a convenient way to observe general word frequency. Lower numbers indicate higher word frequencies in Wikipedia. The 50th and 75th percentile values in the datasets imply that pre-editing induced the avoidance of low-frequency words.

\begin{figure*}[t]
  \centering
  \includegraphics[scale=.4]{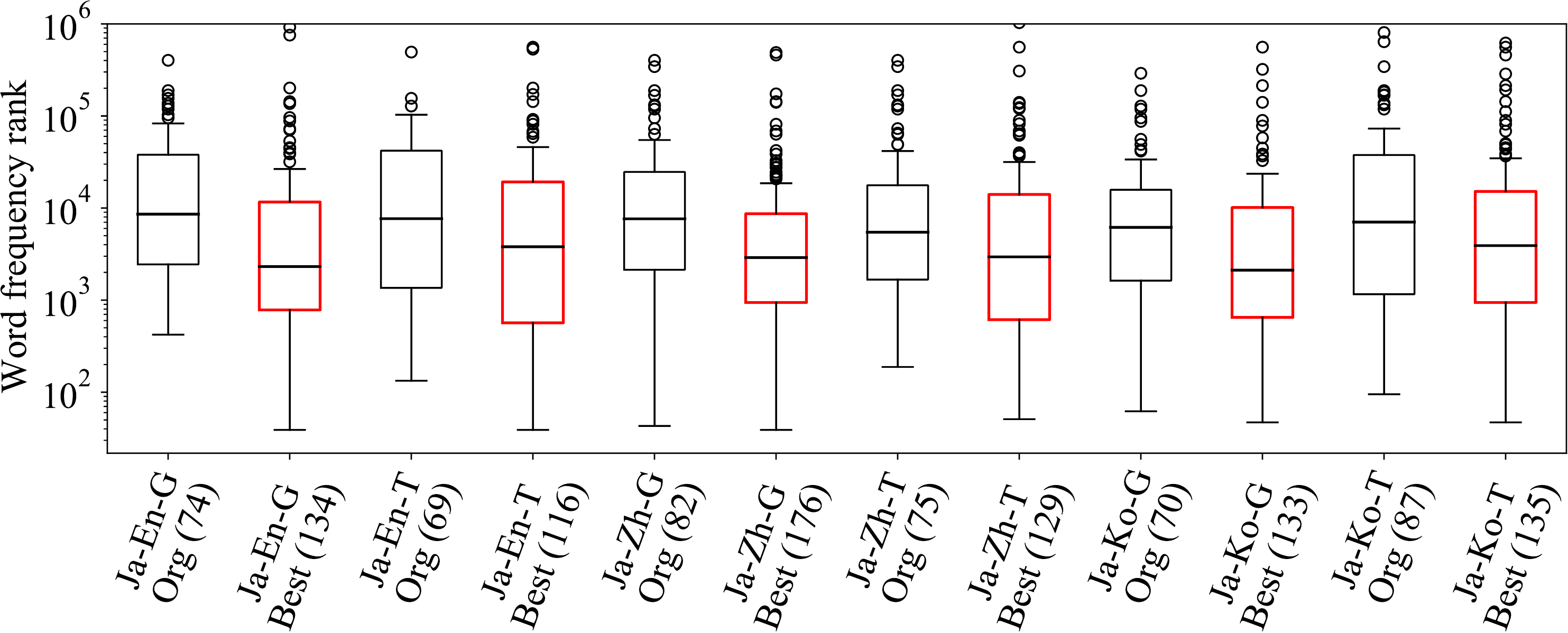}
  \caption{Differences in word frequency rank distribution between the Org-ST and Best-ST (G: Google, T: TexTra). The numbers in parentheses indicate the number of instances, i.e., word types.}
  \label{fig:differences-word-frequency-rank}
\end{figure*}

To further inspect the differences between the Org-ST and the Best-ST, we extracted the word types (a) that appeared only in the Org-ST and (b) that appeared only in the Best-ST. Figure~\ref{fig:differences-word-frequency-rank} illustrates the rank distributions of (a) and (b) for each condition. It is clear that low-frequency words with a frequency rank of around 10,000 decreased in the Best-ST, while words with a frequency rank of around 2,000--4,000 increased in the Best-ST. As \citet{NMT-2017-Koehn} demonstrated, low-frequency words still pose major obstacles for NMT systems. Our results endorse this claim from a different perspective and can provide general strategies for word choice in the pre-editing task.

\hypertarget{sec-diversity}{\section{Diversity of Pre-Editing Operations}}
\label{sec:diversity}

\hypertarget{sub-typology}{\subsection{Typology of Edit Operations}}
\label{sub:typology}

To understand the diversity of edit operations for pre-editing, we manually annotated the collected pre-editing instances in terms of linguistic operations. Given that the Best path contains effective editing operations for improved MT quality, we focused on the pairs of ST versions in the Best path (e.g., the pairs \{1$\to$3, 3$\to$7, 7$\to$8\} in Figure~\ref{fig:best-path}). We randomly selected 10 units for each of the 24 combinations of translation direction, MT system, and text domain, resulting in a total of 961 pre-editing instances. We then excluded 26 instances that could be decomposed into multiple smaller edits\footnote{Only 2.7\% of the edits were not regarded as minimum, which demonstrated satisfactory adherence to our instructions, compared with the implementation by \citet{EAMT-2017-Miyata}, in which 568 pre-editing instances were finally decomposed into 979 instances.} and classified the remaining 935 instances, each of which consists of a minimum edit of ST, based on the typology proposed by \citet{EAMT-2017-Miyata}. Through the classification, we refined the existing typology to consistently accommodate all the instances. 

Table~\ref{tab:typology} presents our typology of editing operations with the number of instances in the different conditions. The typology consists of 39 operation types under 6 major categories, which enables us to grasp the diversity and trends of pre-editing operations. Compared to structural editing, local modifications of words and phrases were frequently used in the Best path. The dominant type is \textbf{C01 (Use of synonymous words)}: content words are replaced by another synonymous word. This operation is important for achieving appropriate word choice in the MT output. \textbf{C07 (Change of content)}, the second dominant type, includes the addition of information that is inferred by human editors based on the intra-sentential context or even external knowledge. For example, a named entity `\textit{Nemuro-sho}' (Nemuro office) was changed into `\textit{Nemuro-keisatsu-sho}' (Nemuro police office) by using the knowledge of the entity. It might be challenging to automate such \textit{creative} operations.

It is also notable that \textbf{S01 (Sentence splitting)} only amounts to 1.5\% of all instances, which supports the observation in \S\hyperlink{sub-structural-characteristics}{4.1} that in general, sentence length was not reduced, and even increased by pre-editing. Among the 14 cases of this type, nine of the split sentences were 60--67 words in length. These results support the empirical observation by \citet{NMT-2017-Koehn} that NMT systems still have difficulty in translating sentences longer than 60 words, and suggest that sentence splitting may only be promising for such very long sentences.

\begin{table*}[t!]
\small
\centering
\begin{tabular}{ll|rrrrrr|r|rrr}
\hline
\multicolumn{1}{c}{\multirow{2}{*}{\textbf{ID}}} & \multicolumn{1}{c|}{\multirow{2}{*}{\textbf{Editing operation type}}} & \multicolumn{2}{c}{\textbf{Ja-En}} & \multicolumn{2}{c}{\textbf{Ja-Zh}} & \multicolumn{2}{c}{\textbf{Ja-Ko}} & \multicolumn{1}{|l|}{\multirow{2}{*}{\textbf{Total}}} & \multicolumn{1}{c}{\multirow{2}{*}{\textbf{Expl.}}} & \multicolumn{1}{c}{\multirow{2}{*}{\textbf{Impl.}}} & \multicolumn{1}{c}{\multirow{2}{*}{\textbf{Pres.}}} \\ \cline{3-8}
\multicolumn{1}{c}{} & \multicolumn{1}{c|}{} & \multicolumn{1}{c}{\textbf{G}} & \multicolumn{1}{c}{\textbf{T}} & \multicolumn{1}{c}{\textbf{G}} & \multicolumn{1}{c}{\textbf{T}} & \multicolumn{1}{c}{\textbf{G}} & \multicolumn{1}{c|}{\textbf{T}} & \multicolumn{1}{l|}{} & \multicolumn{1}{c}{} & \multicolumn{1}{c}{} & \multicolumn{1}{c}{} \\ \Hline
S01 & Sentence splitting & 1 & 0 & 3 & 3 & 4 & 3 & 14 & 0 & 0 & 14 \\
S02 & Structural change & 3 & 5 & 9 & 4 & 4 & 2 & 27 & 8 & 1 & 18 \\
S03 & Use/disuse of topicalisation & 1 & 7 & 4 & 3 & 1 & 3 & 19 & 5 & 2 & 12 \\
S04 & Insertion of subject/object & 2 & 1 & 1 & 3 & 5 & 2 & 14 & 14 & 0 & 0 \\
S05 & Use/disuse of clause-ending noun & 3 & 2 & 2 & 2 & 2 & 1 & 12 & 12 & 0 & 0 \\
S06 & Change of voice & 1 & 3 & 0 & 0 & 0 & 0 & 4 & 2 & 0 & 2 \\
S07 & Other structural changes & 1 & 0 & 2 & 1 & 1 & 0 & 5 & 3 & 0 & 2 \\ \hline
P01 & Insertion/deletion of punctuation & 19 & 16 & 5 & 12 & 9 & 10 & 71 & 0 & 0 & 71 \\
P02 & Use/disuse of chunking marker(s) & 6 & 12 & 2 & 1 & 3 & 4 & 28 & 11 & 8 & 9 \\
P03 & Phrase reordering & 6 & 4 & 7 & 1 & 9 & 4 & 31 & 0 & 0 & 31 \\
P04 & Change of modification & 1 & 3 & 3 & 0 & 0 & 0 & 7 & 0 & 0 & 7 \\
P05 & Change of connective expression & 3 & 18 & 4 & 2 & 10 & 3 & 40 & 24 & 5 & 11 \\
P06 & Change of parallel expression & 3 & 8 & 2 & 8 & 4 & 11 & 36 & 7 & 2 & 27 \\
P07 & Change of apposition expression & 1 & 7 & 2 & 1 & 1 & 4 & 16 & 8 & 4 & 4 \\
P08 & Change of noun/verb phrase & 1 & 3 & 2 & 1 & 3 & 3 & 13 & 9 & 3 & 1 \\
P09 & Use/disuse of compound noun & 1 & 5 & 2 & 2 & 6 & 12 & 28 & 16 & 12 & 0 \\
P10 & Use/disuse of affix & 4 & 4 & 1 & 2 & 3 & 3 & 17 & 1 & 0 & 16 \\
P11 & Change of sahen noun expression & 0 & 1 & 1 & 1 & 2 & 0 & 5 & 1 & 0 & 4 \\
P12 & Change of formal noun expression & 1 & 2 & 2 & 2 & 2 & 0 & 9 & 4 & 0 & 5 \\
P13 & Other phrasal changes & 0 & 1 & 0 & 1 & 2 & 1 & 5 & 4 & 0 & 1 \\ \hline
C01 & Use of synonymous words & 18 & 18 & 19 & 18 & 25 & 20 & 118 & 14 & 10 & 94 \\
C02 & Use/disuse of abbreviation & 2 & 7 & 2 & 2 & 1 & 7 & 21 & 19 & 2 & 0 \\
C03 & Use/disuse of anaphoric expression & 4 & 4 & 2 & 2 & 1 & 1 & 14 & 10 & 2 & 2 \\
C04 & Use/disuse of emphatic expression & 1 & 2 & 2 & 1 & 4 & 1 & 11 & 10 & 1 & 0 \\
C05 & Category indication/suppression & 5 & 3 & 6 & 5 & 4 & 7 & 30 & 29 & 1 & 0 \\
C06 & Explanatory paraphrase & 3 & 4 & 1 & 0 & 1 & 1 & 10 & 0 & 0 & 10 \\
C07 & Change of content & 22 & 20 & 21 & 9 & 14 & 8 & 94 & 57 & 23 & 14 \\ \hline
F01 & Change of particle & 9 & 14 & 4 & 6 & 7 & 7 & 47 & 13 & 5 & 29 \\
F02 & Change of compound particle & 8 & 5 & 5 & 2 & 5 & 6 & 31 & 24 & 2 & 5 \\
F03 & Change of aspect & 1 & 4 & 1 & 0 & 5 & 1 & 12 & 0 & 0 & 12 \\
F04 & Change of tense & 0 & 0 & 1 & 1 & 1 & 1 & 4 & 0 & 0 & 4 \\
F05 & Change of modality & 3 & 1 & 2 & 1 & 3 & 1 & 11 & 5 & 0 & 6 \\
F06 & Use/disuse of honorific expression & 3 & 1 & 1 & 2 & 2 & 1 & 10 & 0 & 0 & 10 \\ \hline
O01 & Japanese orthographical change & 10 & 16 & 9 & 5 & 9 & 12 & 61 & 12 & 4 & 45 \\
O02 & Change of half-/full-width character & 0 & 5 & 3 & 2 & 2 & 4 & 16 & 7 & 1 & 8 \\
O03 & Insertion/deletion/change of symbol & 0 & 2 & 0 & 0 & 0 & 0 & 2 & 0 & 0 & 2 \\
O04 & Other orthographical change & 0 & 1 & 0 & 0 & 3 & 0 & 4 & 0 & 0 & 4 \\ \hline
E01 & Grammatical errors & 0 & 8 & 5 & 2 & 2 & 5 & 22 & -- & -- & -- \\
E02 & Content errors & 5 & 0 & 8 & 1 & 1 & 1 & 16 & -- & -- & -- \\ \hline
\end{tabular}
\caption{Constructed typology of editing operations (G: Google, T: TexTra). The first letter of ID indicates the six major categories (S: Structure, P: Phrase, C: Content word, F: Functional word, O: Orthography, E: Errors casually introduced in the ST). The right three columns provide the frequencies for general informational strategies (Expl.: Explicitation, Impl.: Implicitation, Pres.: Preservation).}
\label{tab:typology}
\end{table*}

\hypertarget{sub-strategies}{\subsection{Strategies for Effective Pre-Editing}}
\label{sub:strategies}

Towards the effective exercise of pre-editing, we further analysed the pre-editing instances in terms of \emph{informational strategies} based on the notion of explicitation/implicitation acknowledged in translation studies \citep{Didier-1958-Vinay,Benjamins-1997-Chesterman,TI-2016-Murtisari}. Following these studies, we broadly defined explicitation as an act of indicating what is implied in the text to clarify its meaning and implicitation as the inverse act of explicitation. We classified all the instances analysed above except for the \textbf{E01} and \textbf{E02} types into three general strategies, namely, explicitation, implicitation, and (information) preservation. The right side of Table~\ref{tab:typology} shows the classification result. The total numbers of instances classified into each strategy were 329, 88, and 480, respectively. Not surprisingly, this indicates that explicitation is an essential strategy for effective pre-editing.

\begin{table*}[t]
\small\centering
\begin{tabular}{llrrlrrlrr}
\hline
 & \multicolumn{1}{l}{} & \multicolumn{2}{c}{\textbf{Ja-En}} &  & \multicolumn{2}{c}{\textbf{Ja-Zh}} &  & \multicolumn{2}{c}{\textbf{Ja-Ko}} \\ \cline{3-4} \cline{6-7} \cline{9-10} 
 & \multicolumn{1}{l}{} & \multicolumn{1}{c}{\textbf{Google}} & \multicolumn{1}{c}{\textbf{TexTra}} &  & \multicolumn{1}{c}{\textbf{Google}} & \multicolumn{1}{c}{\textbf{TexTra}} &  & \multicolumn{1}{c}{\textbf{Google}} & \multicolumn{1}{c}{\textbf{TexTra}} \\ \Hline
\multirow{2}{*}{\textbf{TER}} & \textbf{Pearson's {\boldmath $r$}} & 0.244 & 0.217 &  & 0.144 & 0.204 &  & 0.580 & 0.347 \\
\textbf{} & \textbf{Spearman's {\boldmath $\rho$}} & 0.218 & 0.184 &  & 0.094 & 0.172 &  & 0.574 & 0.248 \\
\hline
\multirow{2}{*}{\textbf{Num. of edits}} & \textbf{Pearson's {\boldmath $r$}}& 0.264 & 0.153 &  & 0.205 & 0.181 &  & 0.465 & 0.212 \\
\multicolumn{1}{c}{\textbf{}} & \textbf{Spearman's {\boldmath $\rho$}} & 0.210 & 0.221 &  & 0.219 & 0.226 &  & 0.449 & 0.245 \\
\hline
\end{tabular}
  \caption{Correlation of the TER and the number of edits between ST and MT.}
\label{tab:correlation}
\end{table*}

We also grouped all the 329 instances of explicitation into the following four subcategories.\footnote{See Appendix~\ref{sec:appendix} for details.}
\begin{description}
  \setlength{\parskip}{0cm} 
  \setlength{\itemsep}{0cm} 
\item[\textbf{Information addition}] is the strategy of adding supplementary information, such as subjects, modality, and explanation, to clarify the content of the ST. 
For example, subjects were sometimes inserted as they tend to be omitted in Japanese sentences. This strategy generally corresponds to operation \textbf{C07 (Change of content)} described earlier.
\item[\textbf{Use of clear relation}] includes structural changes and the use of explicit connective markers to make the relation between words, phrases, and clauses more intelligible. For example, the relation between the subject and object can be clarified by using the nominative case marker `\textit{ga}' in Japanese.
\item[\textbf{Use of narrower sense}] is the strategy of replacing general words with more specific ones. For example, the verb `\textit{dasu},' which has multiple meanings such as `put,' `take,' and `send,' was replaced with the verb `\textit{teishutsusuru},' which has a narrower range of meaning and was correctly translated as `submit.'
\item[\textbf{Normalisation}] includes the use of authorised or standardised expressions, style, and notation. For example, elliptic sentence-ending was completed to construct a normal structure.
\end{description}

These strategies can be used as concise pre-editing principles for human editors and can guide researchers in devising effective tools for pre-editing. We also emphasise that these general informational strategies are not specific to the Japanese language and could be applied to other languages.

\begin{figure*}[t]
  \centering
  \includegraphics[scale=.42]{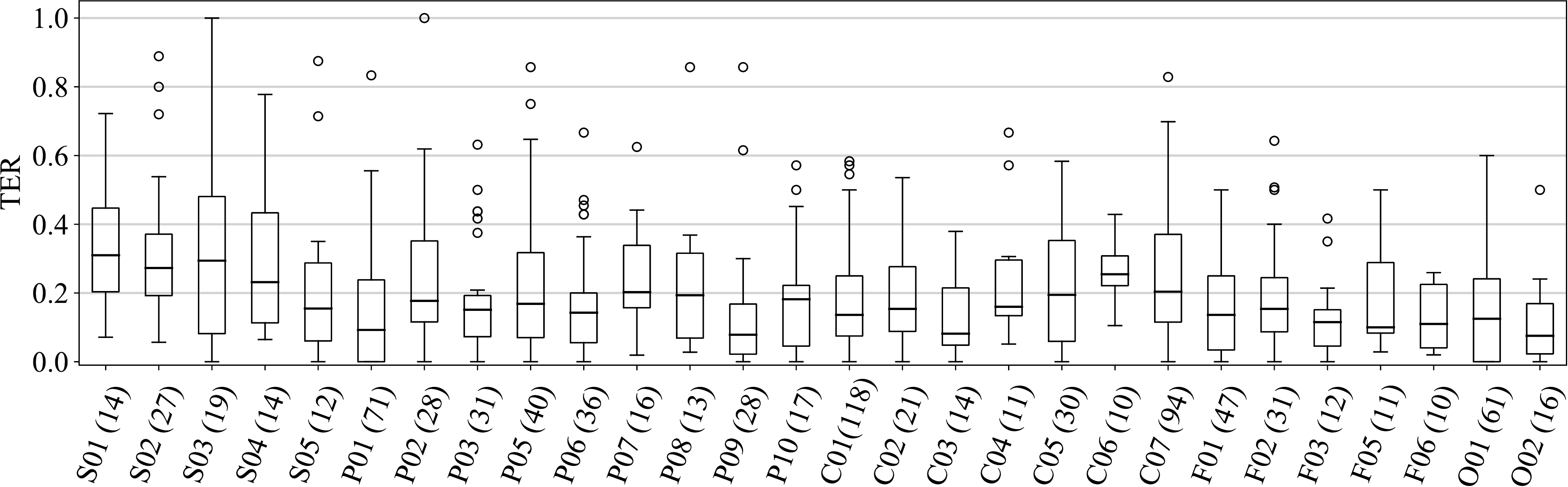}
  \caption{Distribution of the TER for changes in the MT for each operation type with at least 10 instances. The numbers in parentheses indicate the number of instances.}
  \label{fig:operation-ter-boxplot}
\end{figure*}

\hypertarget{sec-impact}{\section{Impact of Pre-Editing on Neural Machine Translation}}
\label{sec:impact}

This section investigates how pre-editing operations affect the NMT output. As indicated in \S\hyperlink{sec-related-work}{2}, NMT systems still lack robustness, and minor modifications of the input would drastically change the output. From the practical viewpoint of deploying pre-editing, predictability is an important object to pursue. Here, we examine the impacts of minimum edits of the ST on the NMT output. To measure the amount of text editing, hereafter, we use the Translation Edit Rate (TER), which is calculated by dividing the number of edits (insertion, deletion, substitution, and shift) required to change a string into the reference string by the average number of reference words \citep{AMTA-2006-Snover}. For any consecutive pair of STs or their corresponding MT outputs, we used the chronologically later version as the reference. For word-level tokenisation, we used MeCab for Japanese, NLTK\footnote{https://www.nltk.org/index.html} for English, jieba\footnote{https://github.com/fxsjy/jieba} for Chinese, and KoNLPy\footnote{https://konlpy.org/en/latest/api/konlpy.tag/\#module-konlpy.tag.\_kkma} for Korean. 

\hypertarget{sub-impact-correlation}{\subsection{Correlation of the Amount of Edits between the ST and MT}}
\label{sub:impact-correlation}

To grasp the general tendency, using all the collected pre-editing instances (see Table~\ref{tab:collected-instances}), we first calculated the correlation coefficients (Pearson's $r$ and Spearman's $\rho$) between the amount of edits (the TER and the number of edits) in the ST and in the MT. More formally, let $\mathtt{ST'}$ be the pre-edited versions of $\mathtt{ST}$. For TER, the correlation is between $\mathtt{TER(ST, ST')}$ and $\mathtt{TER(MT(ST), MT(ST'))}$. For the number of edits, the correlation is between $\mathtt{EditCount(ST, ST')}$ and $\mathtt{EditCount(MT(ST), MT(ST'))}$.

As shown in Table~\ref{tab:correlation}, most coefficients are in the range of 0.15--0.25, suggesting a very weak correlation. This means that the change in NMT output is hardly predictable based on the amount of edits in the ST. For example, the replacement of a single particle in the ST sometimes caused drastic changes of lexical choices in the MT output.

The Japanese-to-Korean translation is an exception; in particular, the correlation coefficients of the TER for the Google NMT system, i.e., 0.580 for Pearson's $r$ and 0.574 for Spearman's $\rho$, indicate a moderate positive relationship between the changes in the ST and those in the MT. This is partly attributable to the fact that the syntactic structures of Japanese and Korean, including the word order and usage of particles, are substantially close. Thus, it is relatively easy to build sufficiently accurate MT systems.

\hypertarget{sub-impact-operation}{\subsection{Impact of Editing Operations on NMT}}
\label{sub:impact-operation}

Finally, using the pre-editing instances in the Best path analysed in \S\hyperlink{sec-diversity}{5}, we further investigated to what extent each type of minimum editing operation affects the MT output. At this stage, we focused on the 28 editing types that have at least 10 instances, considering that it is difficult to derive reliable insights from fewer data.

Figure~\ref{fig:operation-ter-boxplot} presents the distribution of the degree of changes in the MT output when an ST is pre-edited, measured by $\mathtt{TER(MT(ST), MT(ST'))}$. Most of the structural edits (\textbf{S01--S04}) resulted in sizeable changes in the MT. This is reasonable since structural modifications in the ST tended to cause major changes in the MT as well, leading to high TER. In contrast, many of the editing types that include local modifications of functional words and orthographic notations (\textbf{F01--F03, F05, F06, O01, O02}) did not have major impacts on the MT results.

It is worth noticing that \textbf{P03 (Phrase reordering)} did not drastically affect the MT output. In other words, recent NMT systems in practical use manage to retain the phrase-level equivalence even when the position of a phrase is shifted. The influence of \textbf{P02 (Use/disuse of chunking marker(s))} is fairly significant. For human readers, the use of chunking markers, such as double quotes and square brackets, does not greatly affect the sentence parsing, but for NMT, it might seriously impinge on the tokenisation result, eventually leading to a large change in the final output.

\hypertarget{sec-conclusion}{\section{Conclusion and Outlook}}
\label{sec:conclusion}

Towards a better understanding of pre-editing for black-box NMT settings, in this study, we collected instances of manual pre-editing in various conditions and conducted in-depth analyses of the instances. We implemented a human-in-the-loop protocol to incrementally record minimum edits of ST for all combinations of three translation directions, two NMT systems, and four text domains, and obtained a total of 6,652 instances of manual pre-editing. Since more than 95\% of the STs were successfully pre-edited into one that led to a satisfactory MT quality, our collected instances contain empirical, tacit human knowledge on the effective use of black-box NMT systems. We also investigated the collected data from three perspectives: the characteristics of the pre-edited STs, the diversity of pre-editing operations, and the impact of pre-editing operations on the NMT output. The remarkable findings can be summarised as follows:

\begin{itemize} 
  \setlength{\itemsep}{0cm} 
  \item Contrary to the acknowledged practices of pre-editing, the operation of making source sentences shorter and simpler was not frequently observed. Rather, it is more important to make the content, syntactic relations, and word senses clearer and more explicit, even if the ST becomes longer.
  \item As indicated by recent studies, the NMT systems are still sensitive to minor edits in the ST, and are unpredictable in general. However, there are recognisable tendencies in the MT output according to the types of editing operations, such as the relatively small impact of phrase reordering on NMT.
\end{itemize}

In future work, we plan to explore the effective implementation of pre-editing. The findings of this study provide a broad overview of the range of pre-editing operations and their expected benefits, which enables us to find feasible pre-editing solutions in practical use cases of black-box NMT systems. To develop automatic pre-editing tools using a collection of pre-editing instances, we need to handle the data insufficiency issue in machine learning, filling the gap between the training data and targeted black-box MT systems.

Moreover, as our pre-editing instances contain a wide variety of perturbations in the ST, they can also be used to evaluate the robustness of MT systems, which can lead to advances in MT research. We aim to jointly improve the two wheels of translation technology: pre-editing and MT.

\section*{Acknowledgments}

This work was partly supported by JSPS KAKENHI Grant Numbers 19K20628 and 19H05660, and the Research Grant Program of KDDI Foundation, Japan. One of the corpora used in our study was created under a program ``Research and Development of Enhanced Multilingual and Multipurpose Speech Translation System'' of the Ministry of Internal Affairs and Communications, Japan.

\bibliography{eacl2021}

\begin{thebibliography}{36}
\expandafter\ifx\csname natexlab\endcsname\relax\def\natexlab#1{#1}\fi

\bibitem[{Aikawa et~al.(2007)Aikawa, Schwartz, King, Corston-Oliver, and
  Lozano}]{MTS-2007-Aikawa}
Takako Aikawa, Lee Schwartz, Ronit King, Monica Corston-Oliver, and Carmen
  Lozano. 2007.
\newblock Impact of controlled language on translation quality and post-editing
  in a statistical machine translation environment.
\newblock In \emph{Proceedings of the Machine Translation Summit XI}, pages
  1--7, Copenhagen, Denmark.

\bibitem[{Belinkov and Bisk(2018)}]{ICLR-2018-Belinkov}
Yonatan Belinkov and Yonatan Bisk. 2018.
\newblock Synthetic and natural noise both break neural machine translation.
\newblock In \emph{Proceedings of the 6th International Conference on Learning
  Representations (ICLR)}, pages 1--13, Vancouver, Canada.

\bibitem[{Bernth and Gdaniec(2001)}]{MT-2001-Bernth}
Arendse Bernth and Claudia Gdaniec. 2001.
\newblock \href {https://doi.org/10.1023/A:1019867030786} {{MTranslatability}}.
\newblock \emph{Machine Translation}, 16(3):175--218.

\bibitem[{Cheng et~al.(2019)Cheng, Jiang, and Macherey}]{ACL-2019-Cheng}
Yong Cheng, Lu~Jiang, and Wolfgang Macherey. 2019.
\newblock \href {https://www.aclweb.org/anthology/P19-1425/} {Robust neural
  machine translation with doubly adversarial inputs}.
\newblock In \emph{Proceedings of the 57th Annual Meeting of the Association
  for Computational Linguistics (ACL)}, pages 4324--4333. Florence, Italy.

\bibitem[{Chesterman(1997)}]{Benjamins-1997-Chesterman}
Andrew Chesterman. 1997.
\newblock \href {https://doi.org/10.1075/btl.22} {\emph{Memes of Translation}}.
\newblock John Benjamins, Amsterdam.

\bibitem[{Du and Way(2017)}]{PBML-2017-Du}
Jinhua Du and Andy Way. 2017.
\newblock \href {https://doi.org/10.1515/pralin-2017-0018} {Pre-reordering for
  neural machine translation: {Helpful} or harmful?}
\newblock \emph{The Prague Bulletin of Mathematical Linguistics}, 108:171--182.

\bibitem[{Ebrahimi et~al.(2018)Ebrahimi, Lowd, and Dou}]{COLING-2018-Ebrahimi}
Javid Ebrahimi, Daniel Lowd, and Dejing Dou. 2018.
\newblock \href {https://www.aclweb.org/anthology/C18-1055/} {On adversarial
  examples for character-level neural machine translation}.
\newblock In \emph{Proceedings of the 27th International Conference on
  Computational Linguistics (COLING)}, pages 653--663, Santa Fe, New Mexico,
  USA.

\bibitem[{Gulati et~al.(2015)Gulati, Bouillon, Gerlach, Porro, and
  Seretan}]{AIETI-2015-Gulati}
Asheesh Gulati, Pierrette Bouillon, Johanna Gerlach, Victoria Porro, and
  Violeta Seretan. 2015.
\newblock The {ACCEPT Academic Portal}: {A} user-centred online platform for
  pre-editing and post-editing.
\newblock In \emph{Proceedings of the 7th International Conference of the
  Iberian Association of Translation and Interpreting Studies (AIETI)}, Malaga,
  Spain.

\bibitem[{Hartley et~al.(2012)Hartley, Tatsumi, Isahara, Kageura, and
  Miyata}]{EAMT-2012-Hartley}
Anthony Hartley, Midori Tatsumi, Hitoshi Isahara, Kyo Kageura, and Rei Miyata.
  2012.
\newblock Readability and translatability judgments for `{Controlled}
  {Japanese}'.
\newblock In \emph{Proceedings of the 16th Annual Conference of the European
  Association for Machine Translation (EAMT)}, pages 237--244, Trento, Italy.

\bibitem[{Hiraoka and Yamada(2019)}]{MTS-2019-Hiraoka}
Yusuke Hiraoka and Masaru Yamada. 2019.
\newblock \href {https://www.aclweb.org/anthology/W19-6710/} {Pre-editing plus
  neural machine translation for subtitling: {Effective} pre-editing rules for
  subtitling of {TED} talks}.
\newblock In \emph{Proceedings of the Machine Translation Summit XVII}, pages
  64--72, Dublin, Ireland.

\bibitem[{Hoshino et~al.(2015)Hoshino, Miyao, Sudoh, Hayashi, and
  Nagata}]{ACL-2015-Hoshino}
Sho Hoshino, Yusuke Miyao, Katsuhito Sudoh, Katsuhiko Hayashi, and Masaaki
  Nagata. 2015.
\newblock \href {https://www.aclweb.org/anthology/P15-2023/} {Discriminative
  preordering meets {Kendall's} {$\tau$} maximization}.
\newblock In \emph{Proceedings of the 53rd Annual Meeting of the Association
  for Computational Linguistics and the 7th International Joint Conference on
  Natural Language Processing (ACL-IJCNLP)}, pages 139--144, Beijing, China.

\bibitem[{Koehn and Knowles(2017)}]{NMT-2017-Koehn}
Philipp Koehn and Rebecca Knowles. 2017.
\newblock \href {https://www.aclweb.org/anthology/W17-3204/} {Six challenges
  for neural machine translation}.
\newblock In \emph{Proceedings of the 1st Workshop on Neural Machine
  Translation (NMT)}, pages 28--39, Vancouver, Canada.

\bibitem[{Kuhn(2014)}]{CL-2014-Kuhn}
Tobias Kuhn. 2014.
\newblock \href {https://www.aclweb.org/anthology/J14-1005/} {A survey and
  classification of controlled natural languages}.
\newblock \emph{Computational Linguistics}, 40(1):121--170.

\bibitem[{Li et~al.(2007)Li, Li, Zhang, Li, Zhou, and Guan}]{ACL-2007-Li}
Chi-Ho Li, Minghui Li, Dongdong Zhang, Mu~Li, Ming Zhou, and Yi~Guan. 2007.
\newblock \href {https://www.aclweb.org/anthology/P07-1091/} {A probabilistic
  approach to syntax-based reordering for statistical machine translation}.
\newblock In \emph{Proceedings of the 45th Annual Meeting on Association for
  Computational Linguistics (ACL)}, pages 720--727, Prague, Czech Republic.

\bibitem[{Marzouk and Hansen-Schirra(2019)}]{MT-2019-Marzouk}
Shaimaa Marzouk and Silvia Hansen-Schirra. 2019.
\newblock \href {https://doi.org/10.1007/s10590-019-09233-w} {Evaluation of the
  impact of controlled language on neural machine translation compared to other
  {MT} architectures}.
\newblock \emph{Machine Translation}, 33(1-2):179--203.

\bibitem[{Mehta et~al.(2020)Mehta, Azarnoush, Chen, Saluja, Misra, Bihani, and
  Kumar}]{AAAI-2020-Mehta}
Sneha Mehta, Bahareh Azarnoush, Boris Chen, Avneesh Saluja, Vinith Misra,
  Ballav Bihani, and Ritwik Kumar. 2020.
\newblock \href {https://doi.org/10.1609/aaai.v34i05.6369}
  {Simplify-then-translate: {Automatic} preprocessing for black-box
  translation}.
\newblock In \emph{Proceedings of the 34th AAAI Conference on Artificial
  Intelligence (AAAI)}, pages 8488--8495, New York, USA.

\bibitem[{Mirkin et~al.(2013)Mirkin, Venkatapathy, Dymetman, and
  Calapodescu}]{ACL-2013-Mirkin}
Shachar Mirkin, Sriram Venkatapathy, Marc Dymetman, and Ioan Calapodescu. 2013.
\newblock \href {https://www.aclweb.org/anthology/P13-4015/} {{SORT}: {An}
  interactive source-rewriting tool for improved translation}.
\newblock In \emph{Proceedings of the 51st Annual Meeting of the Association
  for Computational Linguistics (ACL), System Demonstrations}, pages 85--90,
  Sofia, Bulgaria.

\bibitem[{Mitamura et~al.(2003)Mitamura, Baker, Nyberg, and
  Svoboda}]{CLAW-2003-Mitamura}
Teruko Mitamura, Kathryn~L. Baker, Eric Nyberg, and David Svoboda. 2003.
\newblock Diagnostics for interactive controlled language checking.
\newblock In \emph{Proceedings of the Joint Conference Combining the 8th
  International Workshop of the European Association for Machine Translation
  and the 4th Controlled Language Applications Workshop (EAMT/CLAW)}, pages
  237--244, Dublin, Ireland.

\bibitem[{Mitamura and Nyberg(2001)}]{NLPRS-2001-Mitamura}
Teruko Mitamura and Eric Nyberg. 2001.
\newblock Automatic rewriting for controlled language translation.
\newblock In \emph{Proceedings of the NLPRS2001 Workshop on Automatic
  Paraphrasing: Theories and Applications}, pages 1--12, Tokyo, Japan.

\bibitem[{Miyata and Fujita(2017)}]{EAMT-2017-Miyata}
Rei Miyata and Atsushi Fujita. 2017.
\newblock Dissecting human pre-editing toward better use of off-the-shelf
  machine translation systems.
\newblock In \emph{Proceedings of the 20th Annual Conference of the European
  Association for Machine Translation (EAMT)}, pages 54--59, Prague, Czech
  Republic.

\bibitem[{Murtisari(2016)}]{TI-2016-Murtisari}
Elisabet~Titik Murtisari. 2016.
\newblock \href {http://www.trans-int.org/index.php/transint/article/view/531}
  {Explicitation in {Translation Studies}: {The} journey of an elusive
  concept}.
\newblock \emph{The International Journal for Translation \& Interpreting
  Research}, 8(2):64--81.

\bibitem[{Niu et~al.(2020)Niu, Mathur, Dinu, and Al-Onaizan}]{ACL-2020-Niu}
Xing Niu, Prashant Mathur, Georgiana Dinu, and Yaser Al-Onaizan. 2020.
\newblock \href {https://www.aclweb.org/anthology/2020.acl-main.755/}
  {Evaluating robustness to input perturbations for neural machine
  translation}.
\newblock In \emph{Proceedings of the 58th Annual Meeting of the Association
  for Computational Linguistics (ACL)}, pages 8538--8544, Online.

\bibitem[{Nyberg et~al.(2003)Nyberg, Mitamura, and
  Huijsen}]{Benjamins-2003-Nyberg}
Eric Nyberg, Teruko Mitamura, and Willem-Olaf Huijsen. 2003.
\newblock Controlled language for authoring and translation.
\newblock In Harold Somers, editor, \emph{Computers and Translation: A
  Translator's Guide}, pages 245--281. John Benjamins, Amsterdam.

\bibitem[{O'Brien(2003)}]{CLAW-2003-O'Brien}
Sharon O'Brien. 2003.
\newblock Controlling controlled {English}: {An} analysis of several controlled
  language rule sets.
\newblock In \emph{Proceedings of the Joint Conference Combining the 8th
  International Workshop of the European Association for Machine Translation
  and the 4th Controlled Language Applications Workshop (EAMT/CLAW)}, pages
  105--114, Dublin, Ireland.

\bibitem[{O'Brien and Roturier(2007)}]{MTS-2007-O'Brien}
Sharon O'Brien and Johann Roturier. 2007.
\newblock How portable are controlled language rules?
\newblock In \emph{Proceedings of the Machine Translation Summit XI}, pages
  345--352, Copenhagen, Denmark.

\bibitem[{Pym(1990)}]{ASLIB-1990-Pym}
Peter Pym. 1990.
\newblock Pre-editing and the use of simplified writing for {MT}.
\newblock In Pamela Mayorcas, editor, \emph{Translating and the Computer 10:
  The Translation Environment 10 Years on}, pages 80--95. Aslib, London.

\bibitem[{Reuther(2003)}]{CLAW-2003-Reuther}
Ursula Reuther. 2003.
\newblock Two in one -- {Can} it work?: {Readability} and translatability by
  means of controlled language.
\newblock In \emph{Proceedings of the Joint Conference Combining the 8th
  International Workshop of the European Association for Machine Translation
  and the 4th Controlled Language Applications Workshop (EAMT/CLAW)}, pages
  124--132, Dublin, Ireland.

\bibitem[{Seretan et~al.(2014)Seretan, Bouillon, and
  Gerlach}]{LREC-2014-Seretan}
Violeta Seretan, Pierrette Bouillon, and Johanna Gerlach. 2014.
\newblock \href {https://www.aclweb.org/anthology/L14-1532/} {A large-scale
  evaluation of pre-editing strategies for improving user-generated content
  translation}.
\newblock In \emph{Proceedings of the 9th International Conference on Language
  Resources and Evaluation (LREC)}, pages 1793--1799, Reykjavik, Iceland.

\bibitem[{Shirai et~al.(1998)Shirai, Ikehara, Yokoo, and
  Ooyama}]{CLAW-1998-Shirai}
Satoshi Shirai, Satoru Ikehara, Akio Yokoo, and Yoshifumi Ooyama. 1998.
\newblock Automatic rewriting method for internal expressions in {Japanese to
  English MT} and its effects.
\newblock In \emph{Proceedings of the 2nd International Workshop on Controlled
  Language Applications (CLAW)}, pages 62--75, Pennsylvania, USA.

\bibitem[{Snover et~al.(2006)Snover, Dorr, Schwartz, Micciulla, and
  Makhoul}]{AMTA-2006-Snover}
Matthew Snover, Bonnie Dorr, Richard Schwartz, Linnea Micciulla, and John
  Makhoul. 2006.
\newblock A study of translation edit rate with targeted human annotation.
\newblock In \emph{Proceedings of the 7th Conference of the Association for
  Machine Translation in the Americas (AMTA)}, pages 223--231, Cambridge,
  Massachusetts, USA.

\bibitem[{Sun et~al.(2010)Sun, O'Brien, O'Hagan, and Hollowood}]{EAMT-2010-Sun}
Yanli Sun, Sharon O'Brien, Minako O'Hagan, and Fred Hollowood. 2010.
\newblock A novel statistical pre-processing model for rule-based machine
  translation system.
\newblock In \emph{Proceedings of the 14th Annual Conference of the European
  Association for Machine Translation (EAMT)}, Saint-Rapha\"{e}l, France.

\bibitem[{Vinay and Darbelnet(1958)}]{Didier-1958-Vinay}
Jean-Paul Vinay and Jean Darbelnet. 1958.
\newblock \emph{Stylistique compar\'ee du fran\c{c}ais et de l'anglais}.
\newblock Didier, Paris, trans. and ed. by J.~C. Sager \& M.-J. Hamel (1995) as
  \textit{Comparative Stylistics of {French} and {English}: {A} Methodology for
  Translation.} John Benjamins, Amsterdam.

\bibitem[{\v{S}tajner and Popovi\'{c}(2018)}]{ATA-2018-Stajner}
Sanja \v{S}tajner and Maja Popovi\'{c}. 2018.
\newblock \href {https://www.aclweb.org/anthology/W18-7006/} {Improving machine
  translation of {English} relative clauses with automatic text
  simplification}.
\newblock In \emph{Proceedings of the 1st Workshop on Automatic Text Adaptation
  (ATA)}, pages 39--48, Tilburg, Netherlands.

\bibitem[{Xia and McCord(2004)}]{COLING-2004-Xia}
Fei Xia and Michael McCord. 2004.
\newblock \href {https://www.aclweb.org/anthology/C04-1073/} {Improving a
  statistical {MT} system with automatically learned rewrite patterns}.
\newblock In \emph{Proceedings of the 20th International Conference on
  Computational Linguistics (COLING)}, pages 508--514, Geneva, Switzerland.

\bibitem[{Yoshimi(2001)}]{MT-2001-Yoshimi}
Takehiko Yoshimi. 2001.
\newblock Improvement of translation quality of {English} newspaper headlines
  by automatic pre-editing.
\newblock \emph{Machine Translation}, 16(4):233--250.

\bibitem[{Zhu(2015)}]{WAT-2015-Zhu}
Zhongyuan Zhu. 2015.
\newblock \href {https://www.aclweb.org/anthology/W15-5007/} {Evaluating neural
  machine translation in {English-Japanese} task}.
\newblock In \emph{Proceedings of the 2nd Workshop on Asian Translation (WAT)},
  pages 61--68, Kyoto, Japan.

\end{thebibliography}
\bibliographystyle{acl_natbib}

\begin{table*}[ht]
  \small
  \centering
  \begin{tabular}{lcp{0.2cm}p{5.3cm}p{4.8cm}}
\hline
\multicolumn{1}{c}{\textbf{Explicitation strategy}} & \textbf{Total} &  & \multicolumn{1}{c}{\textbf{Example of ST pre-editing}} & \multicolumn{1}{c}{\textbf{MT output}} \\ \Hline
\multirow{7}{*}{Information addition} & \multirow{7}{*}{142} & \textbf{} & 12日は台湾の休日のため休場。\newline \textit{12-nichi wa taiwan no kyujitsu no tame kyujo.} & The twelfth is a holiday in Taiwan. \\ \cline{3-5}
 &  & \multirow{4}{*}{\textbf{$\to$}} & 12日は台湾の休日のため\textbf{株式市場は}休場。\newline \textit{12-nichi wa taiwan no kyujitsu no tame \textbf{kabushiki shijo wa} kyujo.} & \textbf{The stock market} was closed on the twelfth due to a holiday in Taiwan. \\ \hline
\multirow{12}{*}{Use of clear relation} & \multirow{12}{*}{103} & \textbf{} & 来院しなくても10日前後で登録のクレジットカードから\textbf{引き落としを行います}。\newline \textit{Raiin-shinakutemo toka zengo de touroku no kurejitto kado kara \textbf{hikiotoshi o okonaimasu}.} & Withdraw from your registered credit card in about 10 days without visiting the hospital. \\ \cline{3-5} 
 &  & \multirow{6}{*}{\textbf{$\to$}} & 来院しなくても10日前後で登録のクレジットカードから\textbf{引き落としが行われます}。　　　　　　　　　　 \textit{Raiin-shinakutemo toka zengo de touroku no kurejitto kado kara \textbf{hikiotoshi ga okonawaremasu}.} & Even if you do not visit the hospital, your credit card will be debited in about 10 days. \\ \hline 
\multirow{5}{*}{Use of narrower sense} & \multirow{5}{*}{54} & \textbf{} & 採尿と採便を\textbf{出して}ください。\newline \textit{Sai-nyo to sai-ben o \textbf{dashite} kudasai.} & Please \textbf{collect} urine and feces. \\ \cline{3-5} 
 &  & \multirow{3}{*}{\textbf{$\to$}} & 採尿と採便を\textbf{提出して}ください。\newline \textit{Sai-nyo to sai-ben o \textbf{teishutsushite} kudasai.} & Please \textbf{submit} urine and stool samples. \\ \hline
\multirow{4}{*}{Normalisation} & \multirow{4}{*}{30} & \textbf{} & 単位は億円。\newline \textit{Tan'i wa oku en.} & \textbf{Figures are in billions of yen}. \\ \cline{3-5} 
 &  & \multirow{2}{*}{\textbf{$\to$}} & 単位は億円\textbf{です}。\newline \textit{Tan'i wa oku en \textbf{desu}.} & \textbf{The unit is 100 million yen}. \\ \hline 
\end{tabular}
\caption{The number of instances and an example of each explicitation strategy for pre-editing ST with MT outputs.}
\label{tab:explicitation}
\end{table*}

\newpage

\renewcommand{\thesection}{A}
\section{Details of Explicitation Strategy}
\label{sec:appendix}

Table~\ref{tab:explicitation} shows the statistics and examples of each subcategory of the explicitation strategy. A total of 329 pre-editing instances of the explicitation strategy can be further classified into four subcategories: information addition, use of clear relation, use of narrower sense, and normalisation.

The example of the information addition illustrates the insertion of a subject `\textit{kabushiki shijo wa}' (`stock market'), which is implicit in the preceding ST. The example of the use of clear relation shows that the relation between the subject and object can be clarified by using the nominative case marker `\textit{ga}' instead of the accusative one `\textit{o}' and accordingly changing the voice of the main clause. As a result, the inappropriate imperative construction `Withdraw from ...' in the MT output is changed to the correct passive construction `will be debited.' In the example of the use of narrower sense, the verb `\textit{dashite},' which has multiple meanings such as `put,' `take,' and `send,' was replaced with the verb `\textit{teishutsushite},' which has a narrower range of meaning and was correctly translated as `submit.' In the example of normalisation, the elliptic sentence-ending was completed with a normal structure `... desu.' This operation led to not only the improvement of the sentence construction, but also the semantic correctness in the MT output (`billions'$\to$`100 million').

\end{document}